\pdfoutput=1

\documentclass[11pt]{article}

\usepackage{acl}

\usepackage{times}
\usepackage{latexsym}
\usepackage{adjustbox}
\usepackage[ruled,vlined]{algorithm2e}
\usepackage[T1]{fontenc}
\usepackage{amsmath}
\usepackage{array}

\usepackage[utf8]{inputenc}
\usepackage{makecell}
\usepackage{microtype}

\usepackage{inconsolata}
\usepackage{hyperref,booktabs}       
\usepackage{url}            
\usepackage{booktabs}       
\usepackage{amsfonts}       
\usepackage{nicefrac}       
\usepackage{microtype}      
\usepackage{xcolor}         
\usepackage{graphicx}
\usepackage{subcaption}
\usepackage{colortbl}
\usepackage{multirow}
\usepackage[skip=2pt]{caption}
\usepackage{wasysym}

\usepackage[textsize=scriptsize]{todonotes}
\usepackage[T1]{fontenc}
\newcommand{\ours}{\textsc{Vis\-Othello}\xspace}

\title{What if Othello-Playing Language Models Could See?}
\author{Xinyi Chen$\thanks{Equal contribution.}^{\spadesuit}$ , Yifei Yuan\footnotemark[1]$^{\heartsuit \clubsuit}$, Jiaang Li$^{\clubsuit}$, Serge Belongie$^{\clubsuit}$, \\ \textbf{Maarten de Rijke}$^{\spadesuit}$, \textbf{Anders Søgaard}$^{\clubsuit}$ \\
$^{\spadesuit}$University of Amsterdam, $^{\heartsuit}$ETH Zürich, 
$^{\clubsuit}$University of Copenhagen \\
\texttt{\{x.chen2, M.deRijke\}@uva.nl, yuanyif@ethz.ch,}\\
\texttt{\{jili, s.belongie, soegaard\}@di.ku.dk}
}

\begin{document}
\maketitle
\begin{abstract}

Language models are often said to face a symbol grounding problem. While some have argued the problem can be solved without resort to other modalities, many have speculated that grounded learning is more efficient. We explore this question in Othello, a simplified, rule-based world that offers a controlled and interpretable testbed for studying world understanding. Building on prior work, we introduce \ours, a multi-modal model trained jointly on move sequences and board images. Using the Othello rule understanding task, we examine whether multi-modal learning provides advantages over text-only approaches. We further evaluate robustness under semantically irrelevant perturbations and analyze the consistency of cross-modal alignment. Our results suggest that multi-modal training not only improves performance and robustness but also promotes convergence toward shared internal representations across different model architectures.

\end{abstract}

\section{Introduction}
Does a language model truly understand what {\em cat} refers to? Of course, none of us fully know what a cat is in an absolute sense, but human language users know enough to use the word appropriately. We can identify cats in images, infer that the furry, mouse-loving pet someone just described is likely a cat, and use the term in context with ease. Whether mono-modal language models can achieve this level of grounding remains an open question. 

This paper does not aim to engage with the discussion over whether symbol grounding is {\em in principle} impossible for mono-modal language models \citep{Mitchell2022TheDO, mollo2023vectorgroundingproblem}. Instead, we focus on the hypothesis that the inclusion of multiple modalities can facilitate more {\em efficient} learning. The question is orthogonal, but entirely consistent with the idea that mono-modal language models can induce (a form of) referential semantics \citep{Sgaard2023-SGAGTV,pmlr-v235-huh24a}.

To test this hypothesis, i.e., to what extent multi-modal language models are more sample-efficient, we turn to the task of learning to play Othello with language models, a domain that offers a well-defined, symbolic environment with clear rules and a compact action space, making it an ideal testbed \cite{li2023emergent,hua2024mothello}. Prior work has used this setup to investigate emergent world representations, training models ranging from small-scale language models~\citep{li2023emergent} to large language models (LLMs)~\cite{yuan2025revisitingothelloworldmodel} to predict the next move from prior moves, with performance evaluated by next legal move accuracy to assess rule learning. A probing classifier is trained to investigate the representations learned for intermediate game states (e.g., my move vs. your move)~\cite{nanda2023emergent}. 
Evidence suggests that language models can learn to track the board state, which potentially forms a rudimentary world model, when trained on large amounts of sequential data. 

\begin{figure*}[t]
\centering
\includegraphics[width=0.93\linewidth]{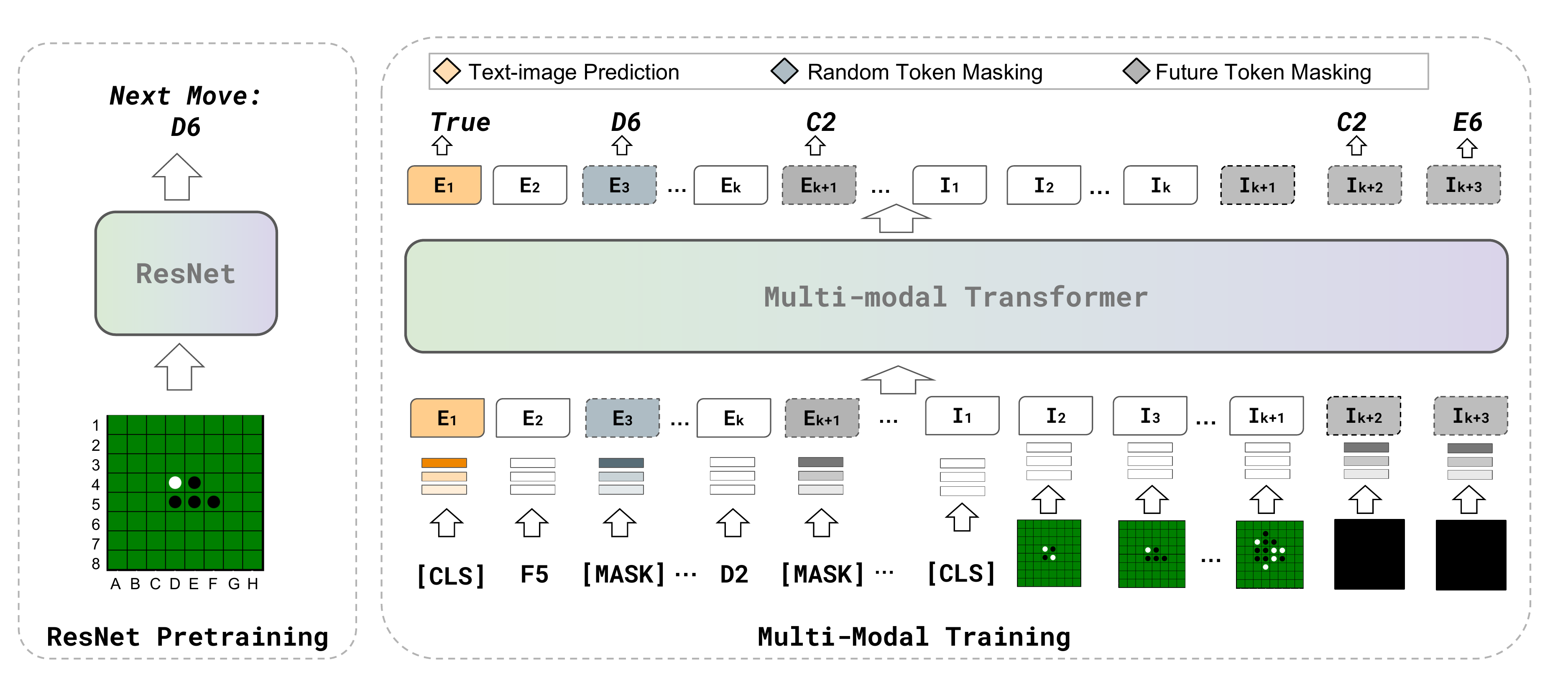}
\caption{Architecture of \ours. The model integrates visual and textual inputs by encoding board images and corresponding move sequences using a Transformer. During pretraining, (i) a ResNet is trained to predict the next move from the current board image; (ii) the multi-modal Transformer is pretrained with three objectives: text-image prediction, random token masking, and future token masking.}
\label{fig:visothello_framwork}
\end{figure*}

While \citet{li2023emergent} showed that text-only models can develop emergent world representations in Othello, our work extends this framework into the multi-modal regime, by introducing \textbf{\ours}, an Othello model trained on sequences of move histories and their corresponding board images (see Figure \ref{fig:visothello_framwork}). 
For each sequence of moves, we generate a corresponding sequence of board state images, with each image depicting the board at a specific time step.
We then apply masking strategies to selected move tokens and train the model to predict the missing steps, using both the move history and associated visual context.

Our main goal is to investigate whether access to visual state information enhances sample efficiency and accelerates learning. We break down the main research questions into several related aspects: 
\begin{center}
\small
\setlength{\tabcolsep}{1.2mm}
\begin{tabular}{ll}
     \toprule 
     {\bf Main question} & Is multi-modal (Othello) learning faster? 
     \\
     \midrule 
     {\bf Sub}$_1$& Is multi-modal learning better? \\
     {\bf Sub}$_2$& Is multi-modal grounding better? \\
     {\bf Sub}$_3$& \makecell[l]{Do multi- and mono-modal models learn \\ aligned representations?} \\
\bottomrule
\end{tabular}
\end{center}
To address {\bf Sub$_1$}, we compare \ours against several baselines on the task of \textit{next move prediction}, where the model predicts the next token given a partial game sequence. We evaluate the performance across varying data scales to assess learning efficiency.
For {\bf Sub}$_2$, we perform a semantically irrelevant perturbation analysis by rotating the board image during inference, assessing whether the models trained on original images remain robust and continue to predict legal moves accurately. Regarding {\bf Sub}$_3$, inspired by \citet{lample2017unsupervised,li2024vision}, we apply two feature alignment techniques to project intermediate representations from different models into a shared vector space and evaluate their similarity.

We show that multi-modal training improves performance and sample efficiency over text-only training. We also observe that multi-modal models exhibit greater robustness to board rotations. Furthermore, through a feature alignment analysis, we find that representational similarity between models increases with more training data---suggesting that, despite differences in architecture and modality, models can converge on shared internal representations.


\paragraph{Contributions.} We are the first to compare the learning curves of mono-modal and multi-modal language models on the task of Othello move prediction and their internal representation learning. We evaluate model robustness by testing invariance to semantically irrelevant perturbations (i.e., board rotations). Additionally, we further study grounding by aligning the internal representations of different models and modalities using supervised and unsupervised methods. Our code is available at \url{https://github.com/shin-ee-chen/multimodal-othello}.

\section{Related Work}
\label{gen_inst}
\subsection{LLMs for Game Sequence Modeling}
Using AI models to play games is not a new concept. Early models, such as AlphaGo, were designed to master gameplay by using predefined game rules and structured environments~\citep{silver2016mastering,silver2017mastering,feng2023chessgpt}.
Recently, modeling games with LLMs and examining their understanding of game dynamics has become a popular research direction in LLM cognitive probing. \citet{li2023emergent} train GPT-2 on synthetically generated Othello games, then use probing techniques to determine whether the model develops internal representations of the game state---effectively inferring a world model. Building on this work, \citet{nanda2023emergent} demonstrate that game-related knowledge is linearly encoded within the model. 
Following this line, research has expanded the scope of world knowledge acquisition in other scenarios with more advanced probing methods~\citep{hao2023reasoning,yun2023emergence,vafa2024evaluating}. For instance, works train similar models with other game datasets, such as chess, maze and checkers,  finding that the same encoding patterns hold in these more complex games~\citep{Karvonen2024EmergentWM,Spies2024TransformersUC,Joshi2024CheckersGPTLW,karvonen2024measuring}. More relevant to our work, \citet{yuan2025revisitingothelloworldmodel} extend the study beyond GPT-2, evaluating state-of-the-art LLMs (e.g., LLaMA-2~\cite{touvron2023llama}, Qwen~\cite{bai2023qwen}) to assess their capacity for structured game knowledge representation. \citet{hua2024mothello} explore this phenomenon in multilingual settings, examining how language models encode and transfer game-related knowledge across different languages. Our work is the first to incorporate visual information in Othello game understanding, providing deeper insights into board state representations.

\subsection{Multi-modal Alignment}
A growing body of research explores cross-modal alignment as a lens to understand the extent to which language models can internalize and generalize knowledge from text-only inputs~\cite{pereira2018toward,caucheteux2022deep,pmlr-v228-li24a,ngo-kim-2024-language}. 
Notably, \citet{merullo2023linearly} demonstrate that visual representations can be effectively projected into the linguistic embedding space using simple linear transformations, revealing a surprising degree of structural compatibility between visual and textual modalities. Building on the theme, \citet{li2024vision} and \citet{pmlr-v235-huh24a} argue that as model capacity increases, representations across modalities tend to converge toward a shared, modality-agnostic statistical structure of the world. Unlike prior work focused on aligning visual and linguistic representations of concrete objects, we extend this to abstract game mechanics, enabling deeper insight into how models understand structured environments from text alone.

\section{Multi-modal Othello Training}
\label{headings}


\subsection{Training Paradigm}
Different from prior works that train Othello models in an autoregressive manner by predicting moves step-by-step, we adopt a \textbf{BERT-style masked language modeling} approach for training \ours. This avoids the computational overhead and complexity of autoregressive generation, enabling efficient bidirectional reasoning over static visual-text inputs without framing the task as video modeling (for a detailed explanation, see Appendix \ref{appendix:modeldesign}).
Specifically, we train \ours based on VisualBERT~\citep{li2019visualbert}. 

\subsection{Input Representation}
\paragraph{Textual input.} Following prior works~\cite{li2023emergent,karvonen2024measuring}, we represent each game as a sequence of moves, where each move at time step $t$ is treated as a token, denoted as $m_t$. Our vocabulary consists of 64 unique tokens, corresponding to the 64 tiles on the board. For example, C4 and E6 correspond to the 27th and 45th token in the vocabulary, respectively. 

\paragraph{Image input.} In addition to the textual input, we provide the model with a sequence of corresponding board images. As demonstrated in Figure~\ref{fig:visothello_framwork}, each image $b_t$ represents the board state after moves $m_1$, $m_2$, \ldots, $m_{t-1}$, and serves as visual context for predicting the next move $m_t$. To extract visual features, considering the differences between Othello board images and object images in ImageNet~\cite{ILSVRC15}, we pretrain an Othello-specific image encoder using a ResNet-18 backbone~\cite{he2016deep}. We then extract visual features using this encoder, resulting in a visual embedding:
\[
\mathbf{v}_t = \phi(\mathbf{b}_t) \in \mathbb{R}^{d_v},
\]
where \( \phi \) denotes the image encoder, and \( d_v \) is the dimensionality of the visual representation. The image embeddings $v_t$ are treated as image tokens to the input of models and are separated from the text tokens by a special token \emph{[SEP]}.



\subsection{\ours Training} \label{subsec:visothello-training}
We train the \ours model using two types of masked language modeling (MLM) strategies to enhance its ability to learn meaningful representations of both textual and visual game sequences. MLM enables the model to develop a deeper understanding of game dynamics.

\paragraph{Random token masking.} Following the training setup of BERT and VisualBERT, we apply random masking to the move sequence with an 80\% probability, masking 15\% of the move tokens at random, while keeping the image tokens fully visible. With the random masking task, the model learns to infer missing information using both modalities, reinforcing cross-modal alignment.

\paragraph{Future token masking.} To align with the next-move prediction setup used in Othello-GPT~\cite{li2023emergent}, we additionally apply future token masking to the game sequence with a 20\% probability. Given a textual move sequence of $m_1$, $m_2$, \ldots, $m_s$, we randomly select a step $t$ ($1 \leq  t \leq  s$) as the prediction target, and then mask all future tokens from $m_t$, which are $m_t$, $m_{t+1}$, \ldots, $m_{s}$. To prevent information leakage, we also mask all the image tokens that contains the future move information, which is $v_{t+1}$, $v_{t+2}$, \ldots, $v_s$. This task encourages the model to rely on previous move sequence rather than future information when predicting the next move. By masking all future tokens and images beyond a randomly selected time step, the model is trained to make predictions in a uni-directional manner, similar to autoregressive models. This setup reduces dependence on bidirectional context and fits better with the next-move prediction setup in the Othello game.

\paragraph{Text-image prediction.}
We also adapt the sentence-image prediction task in original VisualBERT training for the Othello task. For a given sequence of image tokens, we replace the corresponding move sequence to a random sequence at a chance of 50\%. The model is trained to distinguish whether the text and image sequences are from the same game via binary classification. This task helps to train the model better learn implicit alignments between language and vision.

\paragraph{Training objective.} The overall training objective $\mathcal{L}_{\text{total}}$ is defined as the sum of the masked modeling loss and the text-image prediction loss:
\[
\mathcal{L}_{\text{total}} = \mathcal{L}_{\text{mask}} + \mathcal{L}_{\text{ti}},
\]
where the masked modeling loss $\mathcal{L}_{\text{mask}}$ combines two components:
\[
\mathcal{L}_{\text{mask}} = \alpha \cdot \mathcal{L}_{\text{random}} + (1-\alpha) \cdot \mathcal{L}_{\text{future}}.
\]
Here, $\mathcal{L}_{\text{random}}$ is the random token masking loss, $\mathcal{L}_{\text{future}}$ is the future token masking loss, $\mathcal{L}_{\text{ti}}$ is the text-image prediction loss, and $\alpha$ is set to 0.8.

\section{Experiments}\label{others}
In this section, we evaluate whether incorporating visual information improves learning efficiency and grounding in the Othello game setting.  
\subsection{Experimental Setups}
\paragraph{Compared models.}
To better assess the impact of multi-modal learning, we include text-only and vision-only baselines for direct comparison. 


\paragraph{Text-only models.} We evaluate two text-only models with different architectures. (i) Othello-GPT, introduced by \citet{li2023emergent}, is based on GPT-2 and trained autoregressively on Othello move sequences to predict the next move in a purely textual setting. (ii) BERT~\citep{devlin2019bert} is trained using the same language learning objectives as \ours, including both random token masking and future token masking. As BERT serves as the language backbone of our multi-modal model, it provides a strong baseline for isolating the contribution of visual information in learning Othello strategies.

\paragraph{Vision-only models.}
As a vision-only baseline, we train a ResNet-18 model \citep{he2016deep} on board images. Unlike \ours, which processes a sequence of board images and move tokens, the ResNet model is trained to predict the next move based solely on a single board image representing the current game state. It does not observe any move history or future states.




\paragraph{Datasets.}
We collect a total of 25,657 real game records from the \textsc{eOthello} website,\footnote{\url{https://www.eothello.com/}} which serve as the textual sequence inputs for our dataset. For each recorded step in a game, we generate a corresponding image to capture its visual state. As summarized in Table~\ref{tab:dataset_stats}, this process yields approximately 1.56 million images, averaging around 60.8 images per game across all splits. We split the dataset into training (80\%), validation (5\%), and test (15\%) sets, while maintaining a consistent number of images per game across all splits.
\begin{table}[t]
\centering
\resizebox{1\linewidth}{!}{
\begin{tabular}{lccc}
\toprule
\textbf{Split} & \textbf{Games} & \textbf{Images} & \textbf{Avg. per Game} \\
\midrule
Train          & 20,525         & 1,247,852       & 60.8                          \\
Validation     & \phantom{0}1,282          & \phantom{1,0}78,141          & 60.9                          \\
Test           & \phantom{0}3,850          & \phantom{0,}233,975         & 60.8                          \\
\midrule
Total & 25,657 & 1,559,968 & 60.8 \\
\bottomrule
\end{tabular}}
\caption{Dataset statistics. The number of games and images per split. Each game comprises a sequence of steps, with one image per step.}
\label{tab:dataset_stats}
\end{table}
\begin{table*}
\centering
\setlength{\tabcolsep}{12pt}
\small
\begin{tabular}{l cccccc}
\toprule
Train Size & 0 & 1k & 3k & 5k & 10k & 20k \\
\midrule
\rowcolor{gray!30} Othello-GPT  & 7.3$\pm$0.0 & 20.8$\pm$2.8 & 62.1$\pm$15.1 & 66.8$\pm$18.5 & 70.1$\pm$19.8 & 79.7$\pm$2.6 \\
\midrule
BERT-S & 17.1$\pm$2.7 & 89.5$\pm$0.9 & 90.5$\pm$0.5 & 90.8$\pm$0.4 & 91.8$\pm$0.1 & 92.9$\pm$0.1 \\
ResNET-18-S & 21.1$\pm$12.5 & 62.4$\pm$2.3 & 70.9$\pm$3.3 & 74.0$\pm$1.7 & 82.4$\pm$1.4 & 87.2$\pm$2.7 \\
\ours-S & 14.4$\pm$11.3 & \textbf{91.3$\pm$0.5} & \textbf{92.9$\pm$0.9} & \textbf{93.6$\pm$0.3} & \textbf{93.8$\pm$0.3} & 93.8$\pm$0.3 \\
\midrule
BERT-P & 16.7$\pm$2.7 & 88.9$\pm$1.0 & 90.6$\pm$0.7 & 91.4$\pm$0.6 & 91.7$\pm$0.3 & 92.4$\pm$0.5 \\
ResNET-18-P & \textbf{26.3$\pm$21.8} & 73.1$\pm$1.6 & 84.9$\pm$0.3 & 89.2$\pm$1.6 & 91.7$\pm$0.9 & 92.7$\pm$0.5 \\
\ours-P & 25.0$\pm$16.0 & 91.2$\pm$0.5 & 92.3$\pm$0.2 & 93.4$\pm$0.5 & 93.7$\pm$0.3 & \textbf{93.9$\pm$0.2} \\
\bottomrule
\end{tabular}
\caption{Legal move accuracy (\%) for next move prediction in different models across different data sizes. We report mean $\pm$ std over 3 runs, and highlight the best performing model of each training set size in bold. $-P$ indicates the model is pretrained, while $-S$ indicates it is trained from scratch.}
\label{exp1}
\end{table*}

\begin{table*}[h]
\centering
\setlength{\tabcolsep}{12pt}
\small
\begin{adjustbox}{max width=0.99\textwidth}
\begin{tabular}{l cccccc}
\toprule
Train Size & 0 & 1k & 3k & 5k & 10k & 20k \\
\midrule
\rowcolor{gray!30} Othello-GPT & 0.4$\pm$0.0 & 1.9$\pm$0.2 & 16.1$\pm$18.3 & 16.7$\pm$19.6 & 16.8$\pm$17.2 & 27.6$\pm$16.6 \\
\midrule
BERT-S & 1.3$\pm$0.2 & 20.5$\pm$0.2 & 26.2$\pm$1.1 & 29.1$\pm$1.1 & 33.8$\pm$0.3 & \textbf{39.3$\pm$0.3} \\
ResNet-18-S & \textbf{1.6$\pm$0.0} & 10.9$\pm$0.3 & 14.0$\pm$0.3 & 15.4$\pm$0.4 & 18.9$\pm$0.5 & 21.3$\pm$0.2 \\ 
\ours-S & 1.0$\pm$0.7 & \textbf{27.2$\pm$0.3} & \textbf{29.7$\pm$0.6} & \textbf{31.6$\pm$1.1} & 33.5$\pm$0.7 & 36.7$\pm$0.6 \\
\midrule
BERT-P & 1.3$\pm$0.4 & 26.1$\pm$0.2 & 27.7$\pm$0.2 & 31.3$\pm$1.2 & \textbf{34.3$\pm$0.0} & 37.4$\pm$2.1 \\
ResNet-18-P & 1.5$\pm$0.2 & 12.9$\pm$0.5 & 19.5$\pm$0.8 & 22.1$\pm$1.6 & 24.8$\pm$1.7 & 26.7$\pm$0.1 \\
\ours-P & 1.0$\pm$0.4 & 26.1$\pm$0.2 & 29.3$\pm$0.2 & 30.6$\pm$1.2 & 33.4$\pm$0.3 & 35.8$\pm$1.3 \\
\bottomrule
\end{tabular}
\end{adjustbox}
\caption{Exact match accuracy (\%) for next move prediction in different models across different data sizes. For Othello-GPT, BERT, ResNet, and \ours we report mean $\pm$ std over 3 runs.}
\label{exp0}
\end{table*}

\paragraph{Evaluation metrics.} 

When evaluating \ours, we adopt the same setup as the future token masking objective following \citet{li2023emergent}: to predict the legal move at step $m_t$, we mask all future tokens starting from $m_t$, and all image embeddings from $v_{t+1}$ onward, to prevent information leakage from future states.
This setup assesses whether a model can learn the underlying rules of Othello game from sequential move data, in contrast to AlphaZero~\citep{silver2017mastering}, which focus on the winning strategy. Accordingly, we use \textbf{legal move accuracy} as our evaluation metrics. Specifically, we evaluate whether the predicted move $m_t$, given the move history $m_1$, $m_2$, \ldots, $m_{t-1}$,  is valid under Othello’s rules, respectively. 
We also report \textbf{exact match accuracy}, which measures whether the predicted move $m_t$ exactly matches the ground-truth move, reflecting the model's ability to replicate expert gameplay.
\paragraph{Training details.}
To assess the learning efficiency of mono-modal and multi-modal models, we train all models on the full dataset (20k samples) as well as on randomly sampled subsets of 1k, 3k, 5k, and 10k examples. Training and evaluation are conducted with three random seeds (5, 12, and 42). For \ours, we use the best performing image encoder, ResNet-18 pretrained and fine-tuned on the full 20k dataset, for feature extraction.
We also investigate the impact of pretraining by training each model either from scratch or from publicly available pretrained weights.\footnote{For BERT, we use \texttt{google-bert/bert-large-uncased}; the pretrained ResNet-18 is from the HuggingFace \href{https://github.com/huggingface/transformers/blob/main/docs/source/en/model_doc/resnet.md}{transformers} library; the pretrained VisualBERT weights are from the \href{https://github.com/e-bug/volta/blob/main/MODELS.md}{Volta framework}.} For training details, see Appendix~\ref{apx:training_details}. 

\subsection{Experimental Results}
Table \ref{exp1} and \ref{exp0} report the exact match and next legal move prediction accuracy of various models across different dataset sizes. Several key observations emerge from these results. 

\paragraph{Multi-modal learning is more sample efficient.} \ours achieves high accuracy (over 91\% in next legal move prediction) with as few as 1k training examples, while uni-modal models either require more data or fail to reach the same performance ceiling. This suggests that multi-modal learning is more sample-efficient—\ours shows stronger performance at smaller scales than uni-modal baselines. This observation aligns with previous work from \citet{zhuang2024visual}.

\paragraph{Pretraining information is not consistently helpful.}

While pretraining improves performance in some cases, especially in low-data regimes, its effect is not consistent across modalities. With ResNet-18, pretraining is very helpful at small sizes, but its effect decreases as the dataset gets larger. In contrast, for both BERT and \ours, pretraining does not consistently lead to significant gains across training sizes. This observation is consistent with prior findings by \citet{yuan2025revisitingothelloworldmodel}, which suggest that linguistic pretraining may offer limited benefit for structured, rule-based environments such as Othello. 

\subsection{Ablation Study}
\label{appendix:ablation}

To evaluate the contribution of the modified components in \ours relative to the original VisualBERT, we conduct ablation studies focusing on the image encoder and the future token masking strategy.

\paragraph{Image encoder.} We test whether fine-tuning a ResNet model is necessary for extracting image features. For comparison, we consider three alternatives that do not involve task-specific adaptation: (a) a simple pooling projection that downsamples the raw $600 \times 600 \times 3$ image into a 1200-dimensional embedding, (b) an \textit{Area} projection that partitions the image into patches, averages pixel values within each patch, and flattens the resized image into a 1200-dimensional vector, and (c) a ResNet-18 encoder without fine-tuning on Othello images.

\paragraph{Future token masking.} To test the necessity of future token masking (FTM), we train \ours without this component (denoted as \textit{W/O FTM}) and compare performance with the full model.

\paragraph{Results.} As shown in Table~\ref{ablation}, both the choice of image encoder and the use of FTM substantially affect performance. Replacing the fine-tuned ResNet with simple Pooling or Area projections reduces legal move accuracy from 94.03\% to 92.43\% and 91.80\%, respectively, while using an unfine-tuned ResNet yields 92.04\%. These results highlight the benefit of domain-specific adaptation for the visual encoder. More strikingly, removing FTM causes performance to collapse to 62.03\%, underscoring its critical role in aligning the training objective with the causal structure of the game.

\begin{table}[]
\begin{center}
\small
\begin{tabular}{lr}
\toprule
\multicolumn{1}{l}{\bf Method}  &\multicolumn{1}{r} \bf Legal Move Accuracy \\
\midrule
\ours & 94.03\\
\midrule
Pooling   &  92.43\\
Area   &   91.80 \\
W/O FT ResNet     &   92.04\\
\midrule
W/O FTM &  62.03\\
\bottomrule
\end{tabular}
\end{center}
\caption{Ablation results for \ours, using different image encoders (Pooling, Area, ResNet without fine-tuning) and without future token masking (W/O FTM). All results are reported from the best validation checkpoints with training seed 42.}\label{ablation}
\end{table}


\subsection{Probing Internal Representations}
To assess whether \ours learns meaningful internal representations of the board state, we train a linear probe to predict the state of each tile---i.e., whether it is empty, contains the player’s disc, or the opponent’s disc---based on hidden activations after processing a move sequence, following the approach of \citet{nanda2023emergent}. 

\paragraph{Probe training.} The linear probe is trained on 5,000 samples from the training set and evaluated on the same test set used in previous experiments. Due to architectural differences, we restrict this analysis to \ours and the BERT baseline. We evaluate both models after training on 0, 5k, and 20k examples, while keeping the probing setup fixed across all conditions.

\begin{figure}[t]
\centering
\includegraphics[width=1\linewidth]{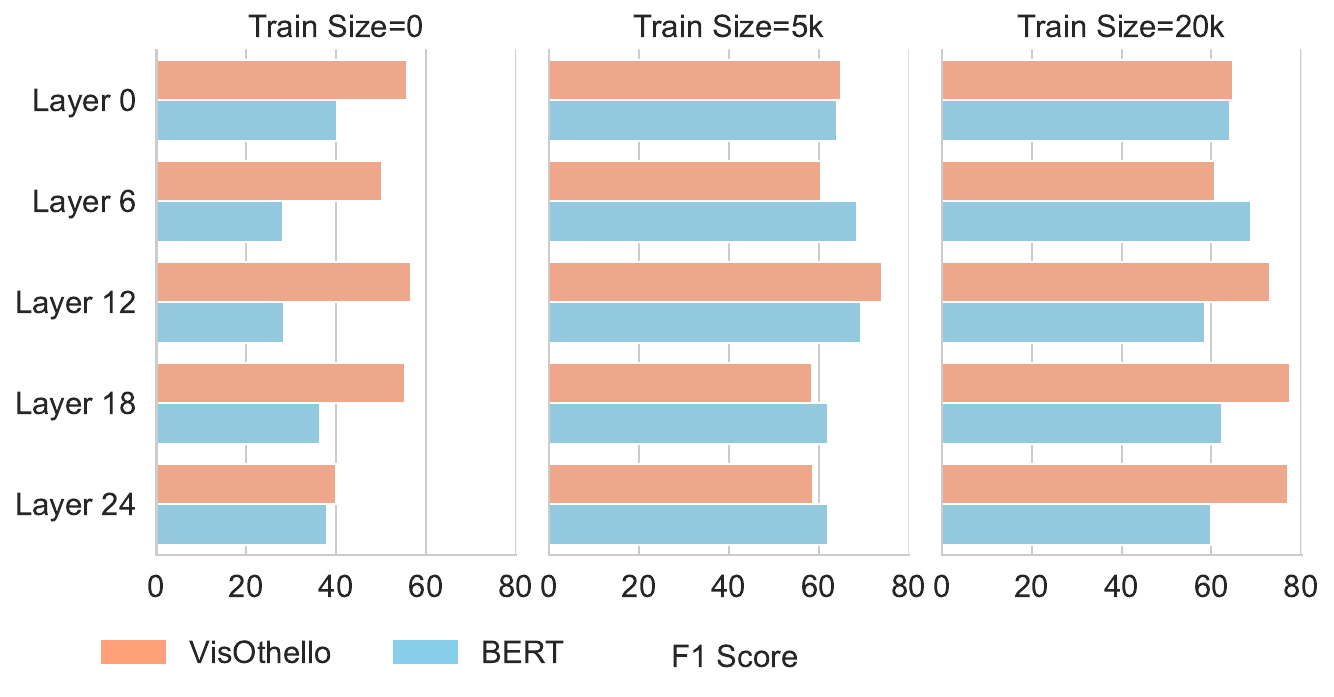}
\caption{Illustration of probing results for BERT and \ours trained with different dataset set sizes.}
\label{fig:probe_result}
\end{figure}

\paragraph{Results.} Figure~\ref{fig:probe_result} shows F1 scores from linear probes trained to predict tile-level board states from selected layers of BERT and \ours. When models are randomly initialized (0 examples), \ours already encodes more board-relevant structure than BERT, achieving substantially higher probe performance in early layers. This may be attributed to the use of a ResNet encoder pretrained on Othello board images, which already encodes useful spatial structure. As training dataset increases, both models improve, but \ours consistently achieves higher scores—especially in deeper layers. After training on 20k examples, \ours reaches 77.55 F1 at Layer 18, compared to 62.28 for BERT. This suggests that \ours learns more accurate internal representations of the board state, benefiting from both multi-modal input and architectural modifications. 


\section{Semantically Irrelevant Perturbation}
 To evaluate model robustness and generalization, we test performance under semantically irrelevant perturbations—input transformations that alter surface form without changing game state.

\subsection{Board Rotation}
\begin{figure}[h]
\centering
\includegraphics[width=1\linewidth]{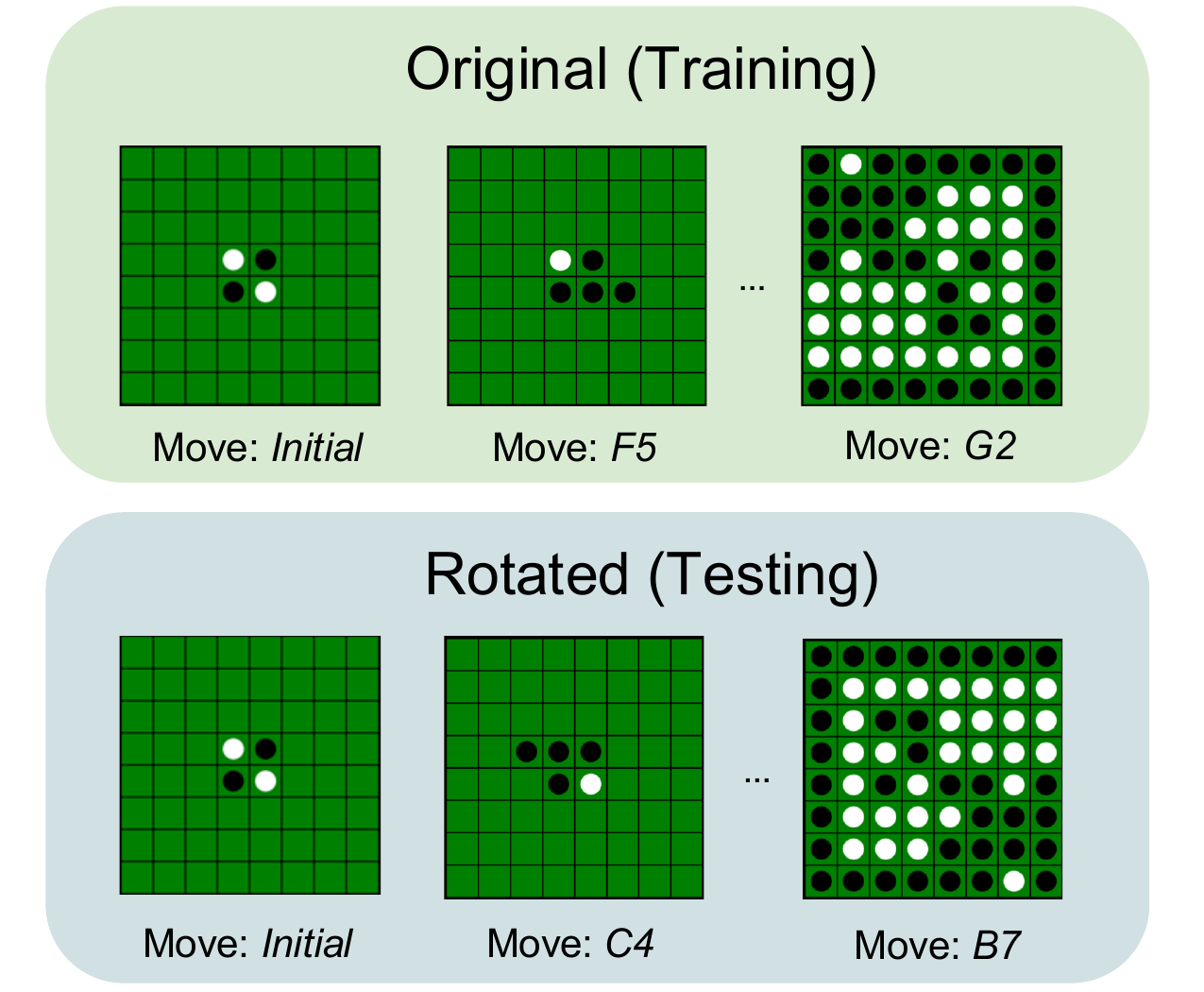}
\caption{Illustration of Rotation $180^{\circ}$. A $180^{\circ}$ rotation preserves game dynamics due to the board’s inherent symmetry and the uniformity of move rules, making such transformations invariant under play.}
\label{fig:rotation_example}
\end{figure}

 We focus on board rotation as a concrete instance, which each test board is rotated 180 degrees. As illustrated in Figure \ref{fig:rotation_example}, this transformation corresponds to a spatial inversion: for image-based models, this involves rotating the game board in the input image; for language-based models, it requires flipping the row and column indices of the move representation (e.g., D3 becomes E6, resulting in a different move token ID).

We apply the rotation \textbf{only at test time}, evaluating models that were trained on the original (unrotated) training data. All models are assessed on their ability to predict both the next legal move and the exact next move, as described in Section~\ref{others}. This setup allows us to examine whether models rely on absolute visual or positional cues, or whether they have learned more abstract, generalizable representations of the board state.

\begin{figure*}[h]
    \centering
    \begin{subfigure}[b]{0.32\textwidth}
        \centering
        \includegraphics[width=\linewidth]{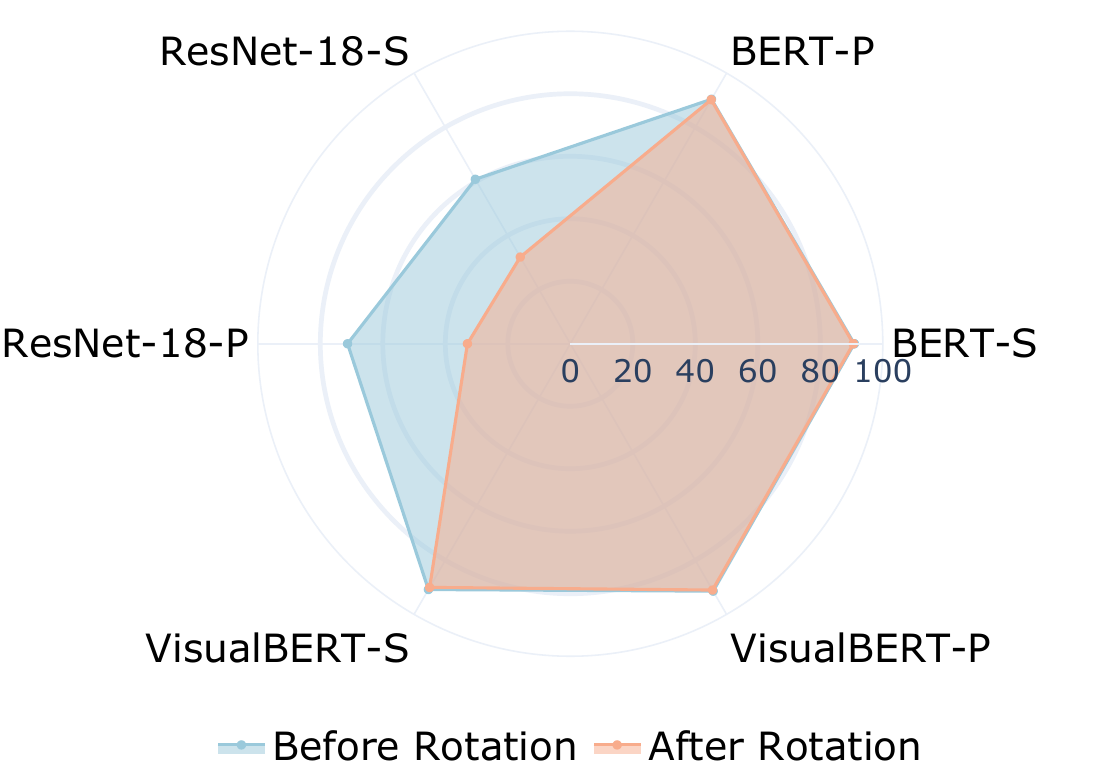}
        \caption{Legal move acc. with 1K samples}
    \end{subfigure}
    \hfill
    \begin{subfigure}[b]{0.32\textwidth}
        \centering
        \includegraphics[width=\linewidth]{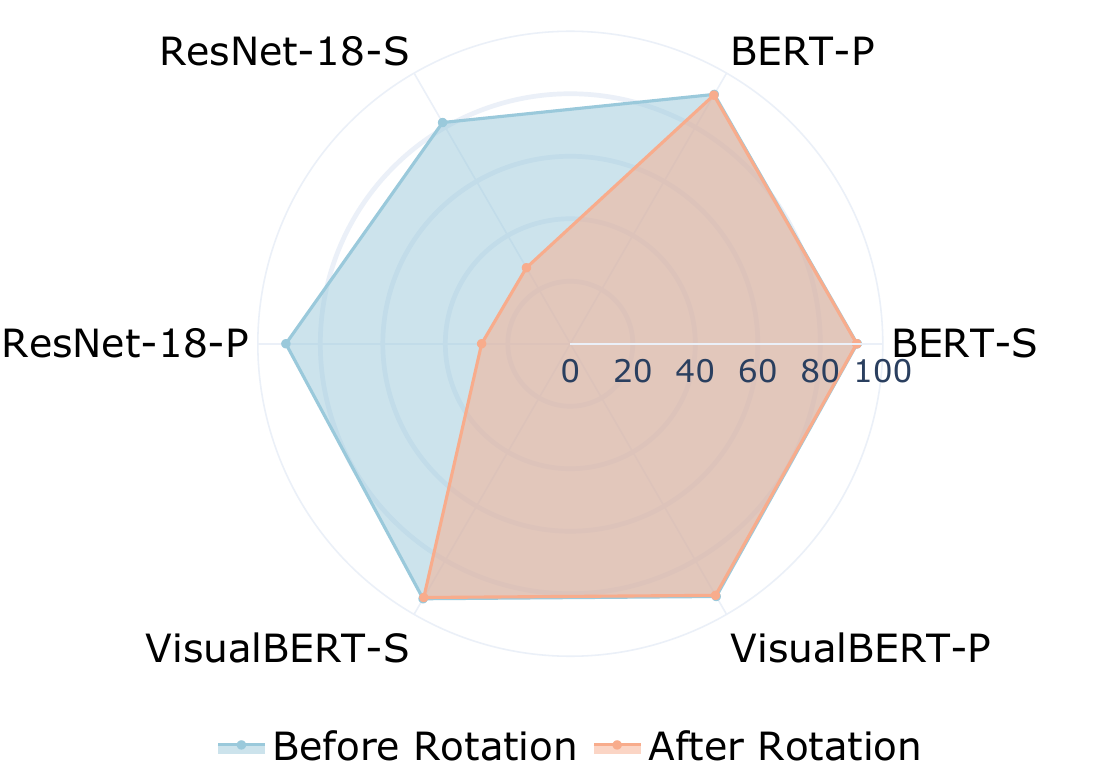}
        \caption{Legal move acc. with 10K samples}
    \end{subfigure}
    \hfill
    \begin{subfigure}[b]{0.32\textwidth}
        \centering
        \includegraphics[width=\linewidth]{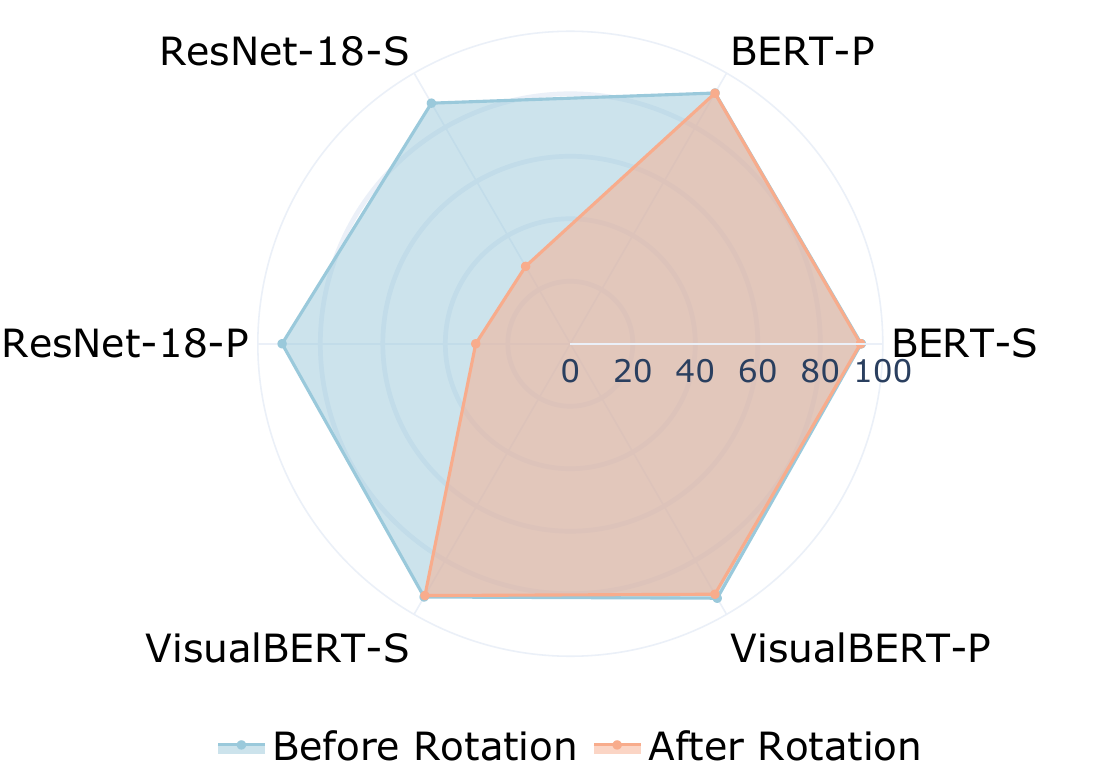}
        \caption{Legal move acc. with 20K samples}
    \end{subfigure}
    
    \begin{subfigure}[b]{0.32\textwidth}
        \centering
        \includegraphics[width=\linewidth]{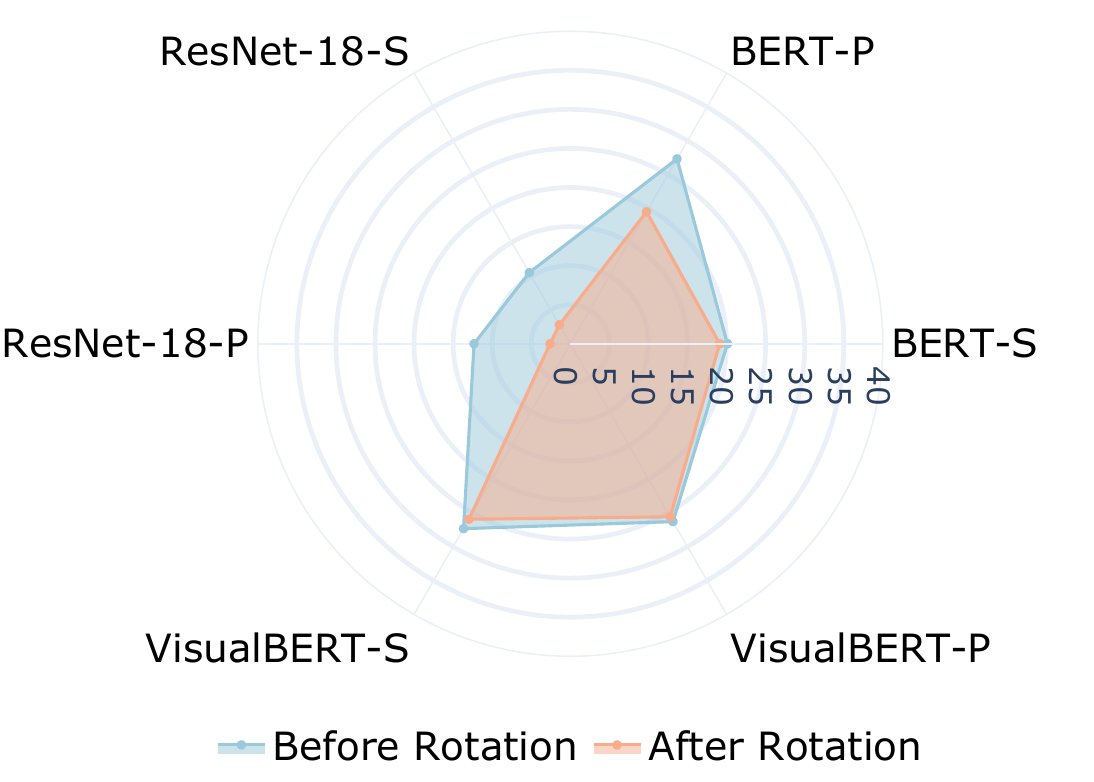}
        \caption{Exact match acc. with 1K samples}
    \end{subfigure}
    \hfill
    \begin{subfigure}[b]{0.32\textwidth}
        \centering
        \includegraphics[width=\linewidth]{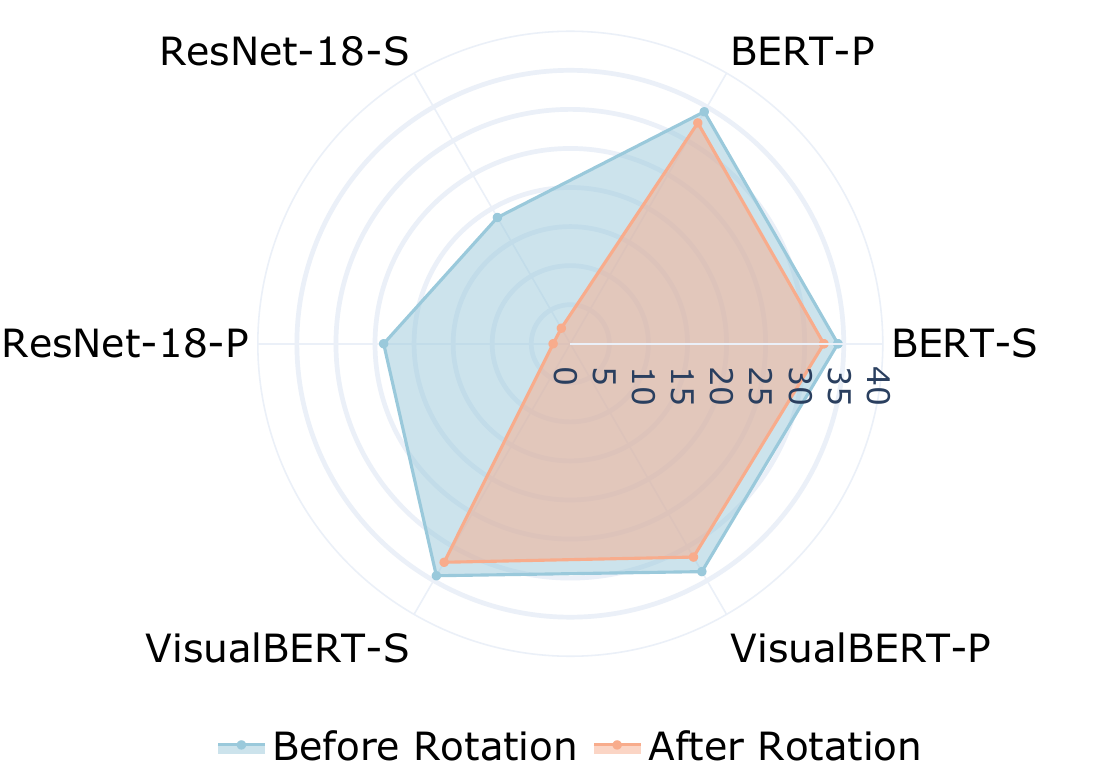}
        \caption{Exact match acc. with 10K samples}
    \end{subfigure}
    \hfill
    \begin{subfigure}[b]{0.32\textwidth}
        \centering
        \includegraphics[width=\linewidth]{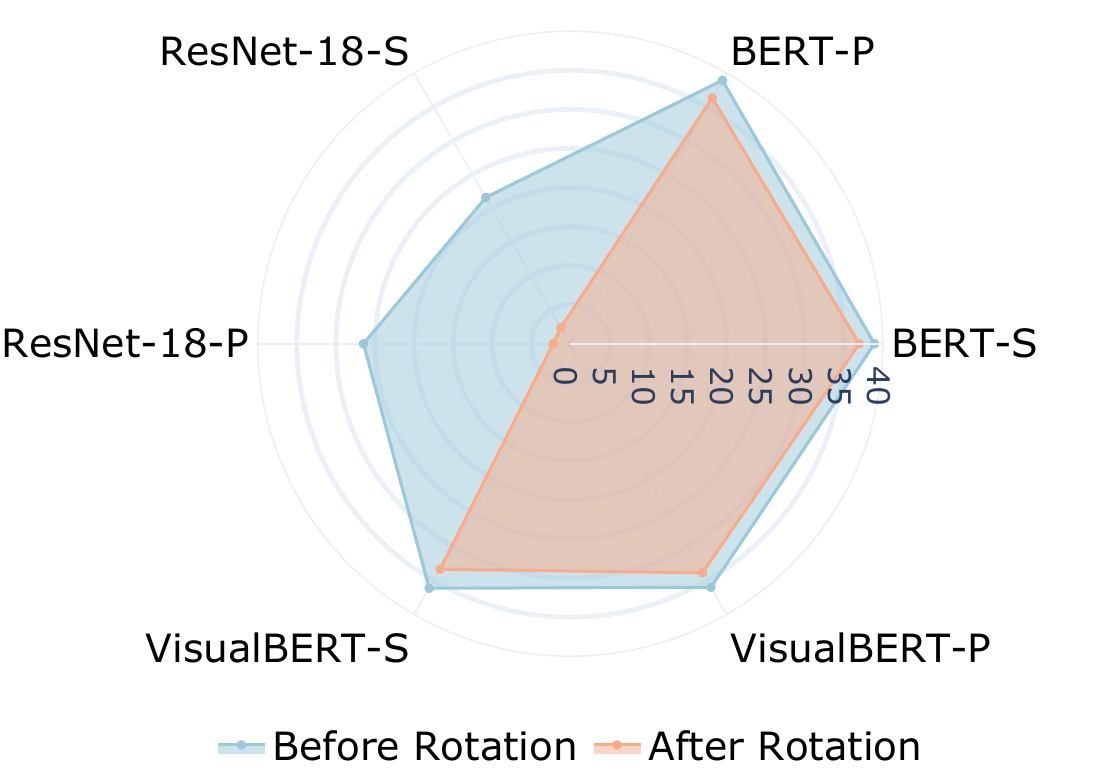}
        \caption{Exact match  acc. with 20K samples}
    \end{subfigure}

    \vspace{0.5cm} 

    \caption{Comparison of models' performance with and without board rotation across different training dataset sizes. The results demonstrate that multi-modal models maintain better performance under rotation compared to purely visual models. $-P$ indicates the model is pretrained, while $-S$ indicates it is trained from scratch.}
    \label{fig:rotation_results}
\end{figure*}

As shown in Figure \ref{fig:rotation_results}, BERT remains relatively robust under board rotation, with accuracy of 90–93\% across all settings. This stability is expected: since BERT operates on symbolic move sequences, board rotation can be handled through a deterministic remapping of move tokens (e.g., D3 becomes E6). In contrast, ResNet-18 suffers a substantial drop in accuracy under rotation, falling to 28–35\% depending on training size and pretraining. This suggests the model fails to learn rotation-invariant representations and relies heavily on absolute spatial patterns. Lacking access to move history or turn information, ResNet depends on ambiguous visual cues that can be easily disrupted—highlighting a key limitation of purely visual models in game playing tasks like Othello.

\ours, which combines ResNet’s image features with BERT’s move sequence encoding, maintains high accuracy after rotation (91–93\%). Compared with ResNet, \ours receives explicit sequence information, including the player turn and previous moves, which helps disambiguate the rotated board. The language modality further guides the interpretation of visual features, enabling the model to maintain stable predictions under spatial transformations. This result illustrates the strength of multi-modal grounding: \textit{by aligning perceptual input with symbolic context, \ours overcomes the spatial brittleness seen in purely visual models}.

\section{Feature Alignment}
We perform representation alignment across models trained on Othello game sequences to assess whether models trained on different modalities (i.e., image and text) learn similar representations. Through this, we investigate whether modality-specific models encode analogous patterns that are fundamental to rule-following gameplay.


\subsection{Alignment Method}
We extract intermediate representations, denoted as $H_i$, from different models for alignment, using the same input sequence, and corresponding board images for multi-modal models. Specifically, we use the features extracted from final hidden layer of both encoder-only models (e.g., BERT, \ours) and decoder-only models (e.g., Othello-GPT). Given the learned representations \( H_1 \) and \( H_2 \) of dimensions \( d_1 \) and \( d_2 \), respectively, extracted from models \( M_1 \) and \( M_2 \) based on the same game sequence input, we first apply PCA to project them into a shared-space of dimension $d = \min(d_1, d_2)$:
\begin{equation}
H_1' = P_{d}(H_1),  H_2' = P_{d}(H_2),
\end{equation}
where $H_1', H_2' \in \mathbb{R}^{d}$ are projected vectors.

Next, we align these representations into a common vector space using the MUSE package,\footnote{\url{https://github.com/facebookresearch/MUSE}} originally developed for mapping multilingual word embeddings into a shared space. The aim is to learn a linear mapping matrix $W$, for each projected representation $H_1'$ and $H_2'$
\begin{equation}
W^* = \mathop{\arg\min}_{W \in \mathcal{M}_i(\mathbb{R})} \| H_i' W - H_j' \|,
\end{equation}
where $i,j\in\{1,2\}$ and $i \neq j$. This denotes learning the optimal linear mapping matrix $W^*$ that aligns representation $H'_i$ to $H'_j$.

\subsection{Alignment Training}
To obtain the optimal mapping matrix, we use both supervised and unsupervised training methods.

\paragraph{Supervised training.} 
We treat representations from different models (e.g., Othello-GPT and \ours) corresponding to the same game sequence as paired training data. For example, given the Othello move sequence input “\textit{F5 F6 E6 F4 C3 D7}”, the pairwise training input $H'_1$ and $H'_2$ correspond to the representations extracted from Othello-GPT and \ours models, respectively, for this exact sequence and the associated images (when applicable). The mapping matrix $W$ is learned and optimized with iterative Procrustes alignment~\citep{gower2004procrustes}, which alternates between solving for the optimal orthogonal transformation and refining the mapping. This process minimizes the distance between the transformed source representations and the target representations, resulting in better alignment across the two vector spaces.

\begin{table*}[]
\centering
\small
\begin{tabular}{ l
    l
    >{\centering\arraybackslash}p{1cm}
    >{\centering\arraybackslash}p{1cm}
    >{\centering\arraybackslash}p{1cm}
    >{\centering\arraybackslash}p{1cm}
    >{\centering\arraybackslash}p{1cm}
    >{\centering\arraybackslash}p{1cm}}
\toprule
\multicolumn{1}{l}{\bf Source}  &\multicolumn{1}{l}{\bf Target} & 0 & 1k & 3k & 5k& 10k & 20k \\
\midrule
BERT & ResNet & 25.37 & 27.39 & 29.48 & 32.07 & 33.34 & 34.25 \\
BERT &  Othello-GPT & \textbf{86.01}& 62.02 & 56.59 & 57.14 & 61.60 & 63.30  \\
BERT & \ours & 83.92 & 61.76 & 54.83 & 53.96 &  55.59 & 57.94\\
Othello-GPT & ResNet &32.16 & 31.79 & 29.95 & 34.33 & 33.74 & 36.41\\
Othello-GPT & \ours & 83.09 & \textbf{77.68} & \textbf{81.62} & \textbf{76.56} & \textbf{82.07} & \textbf{82.35}\\
\ours & ResNet &11.62 & 26.03 & 29.04 & 33.83 & 32.49 & 38.99 \\
\bottomrule
\end{tabular}
\caption{Supervised alignment similarity between target and source models. Highest in bold.}\label{align}
\end{table*}
\begin{table*}[]
\centering
\small
\begin{tabular}{ l
    l
    >{\centering\arraybackslash}p{1cm}
    >{\centering\arraybackslash}p{1cm}
    >{\centering\arraybackslash}p{1cm}
    >{\centering\arraybackslash}p{1cm}
    >{\centering\arraybackslash}p{1cm}
    >{\centering\arraybackslash}p{1cm}}
\toprule
\multicolumn{1}{l}{\bf Source}  &\multicolumn{1}{l}{\bf Target} & 0 & 1k & 3k & 5k& 10k & 20k \\
\midrule
BERT & ResNet& 31.53 & 37.46 & 36.96 & 36.98  & 38.70 & 40.25\\
BERT & Othello-GPT & \textbf{90.38}  & 65.61  & 62.26 & 57.68 & 61.01 & 63.29\\
BERT & \ours & 90.52 & 67.52 & 62.23 & 58.97 & 63.60 & 63.77 \\
 Othello-GPT & ResNet& 33.94 &46.43 & 44.55& 50.15 &46.96& 47.89 \\
 Othello-GPT & \ours & 87.20 & \textbf{80.50} & \textbf{80.53} & \textbf{79.27} & \textbf{85.81} & \textbf{82.46} \\
\ours& ResNet & 23.04 &43.44 &  44.39 & 45.38 & 52.64 & 57.79 \\
\bottomrule
\end{tabular}
\caption{Unsupervised alignment similarity between target and source models. Highest in bold.}\label{align-un}
\end{table*}
\paragraph{Unsupervised training.} We also adopt the unsupervised training approach~\cite{conneau2017word, lample2017unsupervised} with the absence of paired data or predefined anchors to learn the alignment. Given a set of game features $H'$ from both the source and target space, the process begins with adversarial training, where a discriminator is trained to distinguish whether the feature comes from the source or target representation space. Simultaneously, the mapping matrix $W$ is optimized to make this distinction harder, effectively aligning the distributions. Once an initial mapping is obtained, we apply iterative Procrustes refinement, similar to the supervised setting, to improve the alignment. Alignment quality is evaluated and improved using the average cosine similarity between mapped source and target features on the test set.

\subsection{Alignment Training Setups}
To construct the alignment training set, we randomly sample one subsequence from each complete game, resulting in 3,849 input sequences, each paired with the corresponding board state images. We then divide the data into training and testing sets with an 80\%/20\% split, resulting in 3,079 and 770 instances. 
We adopt cosine similarity to measure the alignment quality between representations from different models. After projecting the representations into a shared space, we compute the average pairwise cosine similarity between aligned feature vectors. A higher similarity score indicates better alignment, suggesting that the models, despite being trained on different modalities, capture similar underlying patterns. 
We train the alignment model using a single NVIDIA A100 GPU. All hyperparameters follow the default settings provided by the original MUSE implementation, with no additional tuning. All models in this experiment are trained with a fixed random seed 42.

\subsection{Mapping Result}
Table \ref{align} and \ref{align-un} demonstrate the mapping results under supervised and unsupervised training. We find that the alignment similarity generally improves as the size of the training data increases. This trend suggests that with more data, the models learn richer and shared representations that are easier to align across modalities. Also, despite the difference in training strategy (i.e., autoregressive training and mask language modeling), Othello-GPT and BERT exhibit strong alignment, reflected in their high similarity scores. Surprisingly, Othello-GPT exhibits a strong alignment score with \ours, indicating that despite differences in architecture and training modalities, the two models learn remarkably similar representations. This suggests that the underlying patterns essential for Othello gameplay are captured consistently across both language-based and multi-modal models. Such alignment highlights the potential for cross-modal knowledge transfer and opens avenues for further exploration of unified representations in complex tasks.


\section{Conclusion}

We studied the task of learning to play Othello and extended it to a multi-modal setting by introducing \ours. Our experiments examined whether access to visual state information improves sample efficiency and accelerates learning, comparing \ours against text-only and vision-only baselines. To further assess the benefits of multi-modal grounding, we introduce a board rotation perturbation and conduct feature alignment analysis to evaluate whether the models learn more robust and aligned representations. Our findings suggest that grounding language models with visual input leads to more efficient and stable learning. Beyond Othello, our framework provides a controlled testbed for the analysis of grounded representations and has the potential to extend to other model architectures, tasks, and modalities (see Appendix \ref{appendix:potentialimpacts} for further discussion).

\section*{Limitations}
A notable limitation of this work is that we are not able to compare \ours with autoregressive multi-modal large language models (MLLMs) due to fundamental differences in training paradigms. Autoregressive MLLMs treat images as part of a sequential token stream, effectively converting static visual-text inputs into video modeling tasks, which significantly increases computational complexity and alters the problem structure. In contrast, our model uses masked language modeling (MLM) to enable efficient bidirectional reasoning over static data, making direct comparison with autoregressive MLLMs infeasible without substantial task reformulation.

Moreover, we do not include comparisons with large-scale text-only language models, as these have been thoroughly investigated in prior work~\cite{yuan2025revisitingothelloworldmodel}. Given that pretraining on language alone does not necessarily enhance understanding of the structured reasoning inherent in Othello, scaling up to such models and benchmarking against them is not currently a priority. Instead, our use of lightweight language models offers a practical and efficient probe into how much language pretraining contributes to this domain.


\section*{Ethics Statement}
We ensure that all datasets used in this work are publicly available and released under appropriate open-source licenses. No personal information about players or tournaments is included or revealed. Additionally, all corresponding images used in our experiments are synthetically generated, and do not depict real individuals or contain sensitive content.

\section*{Acknowledgments}
We thank the reviewers and area chair for their suggestions to incorporate the world model, which helped clarify our contributions better. We are also grateful to the members of the CoAStaL group at the University of Copenhagen and the IRLab at the University of Amsterdam for their valuable feedback on the experiments. Xinyi Chen is funded by the project LESSEN (NWA.1389.20.183) of the research program NWA-ORC 2020/21, which is (partly) financed by the Dutch Research Council (NWO), as well as by travel support from ELIAS (GA No.101120237) through the ELLIS program. Serge Belongie and Jiaang Li are supported by the Pioneer Centre for AI, DNRF grant number P1.
Maarten de Rijke was supported by the Dutch Research Council (NWO), under project numbers 024.004.022, NWA.1389.20.\-183, and KICH3.LTP.20.006, and the European Union under grant agreements No. 101070212 (FINDHR) and No. 101201510 (UNITE).
Views and opinions expressed are those of the author(s) only and do not necessarily reflect those of their respective employers, funders and/or granting authorities.


\bibliography{anthology,custom}

\newpage
\appendix
\section{Potential Impacts}
\label{appendix:potentialimpacts}
Our multi-modal Othello framework demonstrates how integrating visual and textual modalities can enhance structured reasoning in environments with strict rule-based dynamics. Beyond board games, this approach offers insights into multi-modal learning for tasks requiring spatial-temporal understanding, such as strategy modeling, robotics, and educational AI systems. By disentangling perceptual and symbolic reasoning, it also serves as a testbed for evaluating how models learn abstract rules from multi-modal input, potentially informing the design of more robust, interpretable, and generalizable multi-modal AI systems. Future work may generalize these insights to more complex domains and explore the role of other modalities, such as spatial or tactile input, in supporting the emergence of grounded representations.

While Othello is a harmless testbed, our methods for aligning multimodal features could, in principle, be adapted to sensitive domains. Our findings on data efficiency may also lower the barrier to training multimodal agents in low-resource settings. We emphasize that our work is intended solely for research on interpretability and grounding.

\section{Model Design Motivation}
\label{appendix:modeldesign}
We use a masked language model (MLM) rather than an autoregressive multi-modal large language model (MLLM) for two main reasons detailed below.

\paragraph{Autoregressive training paradigm is not well-suited for our task setup.} Othello involves dynamic visual changes, as discs flip after each move (Figure~\ref{fig:visothello_framwork}). Understanding the current board state requires access to the complete move history, as it cannot be inferred from a single image alone—even for human players. Thus, the input must consist of a sequence of move tokens paired with the corresponding sequence of board states. This token-aligned multimodal sequence deviates significantly from standard MLLM training paradigms, which are typically designed for single image–text pairs or interleaved inputs without sequential dependencies. A more suitable framing is to model the game as a video sequence, with each board as a frame. However, feeding full sequences into current MLLMs introduces the risk of information leakage from future states and would require specialized causal multimodal masking, implying non-trivial architectural and training modifications.

\paragraph{Autoregressive training incurs extremely high resource costs.} Even if the technical challenges above were addressed, autoregressive training would remain computationally demanding: modeling an $n$-step game requires $n$ forward passes with progressively longer input sequences, whereas our MLM objective learns from the entire game in a single pass with partial masking. Our experiments were conducted on a single A100 GPU (40GB) with a dataset of approximately 20k Othello games, a scale that makes MLLM training infeasible.

Since the goal of this paper is to investigate the role of images in model understanding, we adopt an MLM-based approach with VisualBERT rather than MLLMs. This choice provides a lightweight framework for probing and analyzing the representations learned in multimodal Othello training.



\section{Model Training Details}
All models are trained for up to 1000 epochs, with validation performed every 10 epochs. We apply early stopping with a patience of 5 validation steps, and retain the checkpoint with the highest validation accuracy for final evaluation. Training is conducted on a single NVIDIA A100-40GB GPU. BERT and \ours are trained with a batch size of 128 and a learning rate of 1e-4, while ResNet is trained with a batch size of 512 using the same learning rate.

\label{apx:training_details}

\section{Model Sizes And Compute Resources}
We report the parameter sizes and compute resources for all models used in our experiments. \ours, based on VisualBERT-base, has about 112M parameters. BERT (25 layers, hidden size 768) has about 177M parameters. Othello-GPT, based on GPT-2 Medium, contains 345M parameters. ResNet-18 has about 11.7M parameters. All models were trained on a single NVIDIA A100-40GB GPU. Training on 20k tasks required approximately 10 hours for \ours, 0.5 hours for ResNet-18, and 3.5 hours for BERT




\end{document}